\documentclass[letterpaper, conference, 10pt]{IEEEconf}
\IEEEoverridecommandlockouts

\usepackage{times}

\usepackage{amsmath,amssymb,amsfonts,amsthm}
\usepackage{mathtools, nccmath}
\usepackage{algorithm}
\usepackage{algpseudocode}
\usepackage{graphicx}
\usepackage{textcomp}
\usepackage{soul}
\usepackage{comment}
\usepackage{bm}
\usepackage{array, makecell}
\usepackage{dsfont}
\usepackage{ulem}
\usepackage{optidef}
\usepackage{booktabs}
\usepackage{multicol}
\usepackage{subfig}
\usepackage[dvipsnames]{xcolor}

\usepackage[font=footnotesize]{caption}
\usepackage[bookmarks=true]{hyperref}

\newtheorem{Pro}{Problem}

\newtheorem{Lem}{Lemma}

\newtheorem{Def}{Definition}

\hypersetup{
    colorlinks=true,
    linkcolor=blue,
    filecolor=black,      
    urlcolor= black,
    citecolor=blue,
    pdftitle={Overleaf Example},
    pdfpagemode=FullScreen,
    }

\begin{document}

\title{\LARGE \textbf{Energy Efficient Planning \\  for Repetitive Heterogeneous Tasks in Precision Agriculture}}
\author{Shuangyu Xie$^{1}$, Ken Goldberg$^{1,2}$, and Dezhen Song$^{3,\dagger}$
    \thanks{$^{1}$Department of Electrical Engineering and Computer Science}%
    \thanks{$^{2}$Department of Industrial Engineering and Operations Research}%
    \thanks{$^{1,2}$University of California, Berkeley, CA, USA }
    \thanks{$^{3}$Department of Robotics, Mohamed Bin Zayed University of Artificial Intelligence (MBZUAI), Abu Dhabi, UAE.}%
    \thanks{ $^\dagger$ For correspondence and questions: \texttt{dezhen.song@mbzuai.ac.ae}.
    }
}

\maketitle

\begin{abstract} 
Robotic weed removal in precision agriculture introduces a repetitive heterogeneous task planning (RHTP) challenge for a mobile manipulator. RHTP has two unique characteristics: 1) an observe-first-and-manipulate-later (OFML) temporal constraint that forces a unique ordering of two different tasks for each target and 2) energy savings from efficient task collocation to minimize unnecessary movements. RHTP can be framed as a stochastic renewal process. According to the Renewal Reward Theorem, the expected energy usage per task cycle is the long-run average. Traditional task and motion planning focuses on feasibility rather than optimality due to the unknown object and obstacle position prior to execution. However, the known target/obstacle distribution in precision agriculture allows minimizing the expected energy usage. For each instance in this renewal process, we first compute task space partition, a novel data structure that computes all possibilities of task multiplexing and its probabilities with robot reachability. Then we propose a region-based set-coverage problem to formulate the RHTP as a mixed-integer nonlinear programming. We have implemented and solved RHTP using Branch-and-Bound solver. Compared to a baseline in simulations based on real field data, the results suggest a significant improvement in path length, number of robot stops, overall energy usage, and number of replans.
\end{abstract}




\section{Introduction}
\begin{figure}[t!]
\centering
\includegraphics[width = 2.7in]{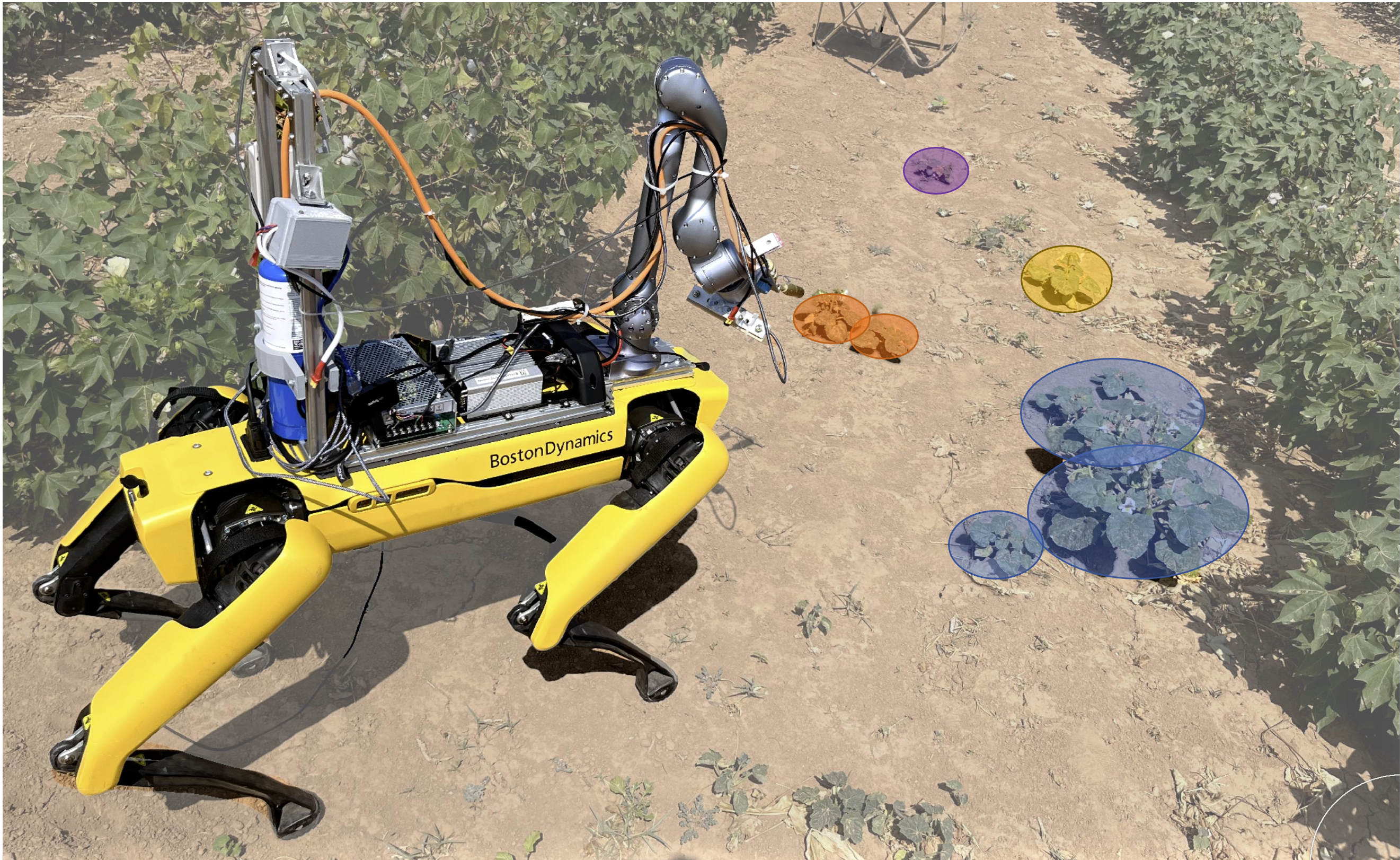}
     \caption{A weed removal robot operating in a cotton field, where the target weeds are highlighted as colored regions. Weed regions sharing the same color belong to the same cluster. The RHTP algorithm optimizes observation and weed flaming task. }
     \label{fig:scene}
     \vspace*{-0.35in}
\end{figure}

Deploying robotic systems in precision agriculture reduces labor requirements and avoids over-application of chemicals. We present a robotic weed removal planning algorithm to guide a mobile manipulator (Fig.~\ref{fig:scene}) that traverses a field to precisely burn key growth spots on each weed using a flame torch, which is environmentally friendly and gaining popularity in applications~\cite{ datta2013flaming, wang2024preciseroboticweedflaming}. Such tasks are often repetitive in nature, and energy usage can be reduced from task collocation if planned efficiently. As a repetitive heterogeneous task planning (RHTP) problem, the weed removal process is a renewal-reward process. According to the Renewal Reward Theorem \cite{TEXTBOOK-sheldon-Ross2024}, the expected energy usage per task cycle is the long-run average.   Consequently, it is imperative to minimize the expected energy usage per cycle when determining the robot action sequence and paths. Traditional integrated Task and Motion Planning (TAMP) \cite{garrett2021integrated} typically addresses single-instance cases, prioritizing solution feasibility over optimality due to uncertain object and obstacle positions. However, precision agriculture often provides known spatial distributions of targets, enabling optimization of expected energy usage for repetitive tasks.


Robotic weed flaming applications also pose new challenges. Targeting peripheral leaves or branches may not effectively suppress or eliminate weeds \cite{datta2013flaming}. Since real-time high-resolution field maps are often inaccessible, robots must conduct close-range inspections to precisely locate each weed's growth point for effective flaming. This approach introduces two key task constraints: i) each task requires the robot to first observe the target, revealing its precise position before executing the manipulation step, and ii) multiple clustered targets reduce robot base movements to save energy and increase operational speed if planned carefully. These factors result in heterogeneous tasks with specific sequencing requirements and opportunities for optimized movement. Similar properties are common in various precision agriculture applications, including harvesting and spraying.


We solve the RHTP problem by minimizing the expected energy usage per cycle. We propose a partition-based task space approach by computing all possibilities/probabilities of task multiplexing while considering the robot's reachability constraint. This transforms the original problem into a region-based optimization problem that focuses on spatial task multiplexing and sequencing. We implemented and tested our proposed algorithm in simulations using data collected from a cotton field with our custom platform (Fig.~\ref{fig:scene}). The results show that our RHTP algorithm significantly reduces the path length, the number of robot stops, the energy cost, and the number of replans compared to its counterpart. Notably, these improvements become more substantial as the target density increases, which is desirable for practical agricultural scenarios.

\begin{table*}[htbp!]
\small
    \centering
        \caption{	 LHP \& complex planning problem comparison. * means approximation to the optimal.}
        \resizebox{\textwidth}{!}{%
    \begin{tabular}{l cccc}\toprule[0.8pt]

         \textbf{Problem \& Algorithm} &  \textbf{Task} & \textbf{Platform}   & \textbf{Scene Knowledge} & \textbf{Metric \& Optimality} \\ \midrule \addlinespace[.1em]
         Sensor Planning \cite{agriculture_UAV_UGV}               & N & UAV+UGV             & Complete               &  Optimal trajectory*  \\ \addlinespace[.1em]
         Tabletop Object Rearrangement \cite{gao2023orla}         & M & Mobile Manipulator  & Complete              &  Optimal p\&p sequence \\ \addlinespace[.1em]
         Blindfolded Traveler Problem \cite{saund2019blindfolded} & PO + N & Mobile robot        & Partial, Incremental &   Optimal trajectory*   \\ \addlinespace[.1em] 
         TAMP w. Partial Observation \cite{garrett2020online}     & PO+N+M & Mobile Manipulator  &  Partial, Incremental  & \makecell{ Feasible task plan, \\ optimal state trajectory*}\\ \addlinespace[.1em]
         \textbf{RHTP (ours)}                                   & AP+N+M & Mobile Manipulator  &  Partial, Incremental & Long-run optimal task\&state*  \\ \hline
  
    \end{tabular}}

    \label{tab:alg_compare}
\vspace{-0.2in}

\end{table*}

\section{Related Work}

We view this RHTP problem as a special case of the long-horizon planning (LHP) problem~\cite{Long-horizon-planning}.  LHP is a challenging planning problem that generates robot behaviors or trajectories to achieve complex tasks or navigation objectives in an environment of uncertainty \cite{kurniawati2011motion}. This LHP problem is related to informative path planning \cite{agriculture_UAV_UGV}, multi-object manipulation \cite{GaoFenHuaYu23IJRR,Manipulating-multi-object-survey}, navigation with uncertainty \cite{saund2019blindfolded,GP-MPPI}, and the TAMP framework \cite{wolfe2010combined,garrett2021integrated,kaelbling2013integrated}. We compare our work with closely related works and summarize the comparison in Table~\ref{tab:alg_compare} as detailed below.


\textbf{Tasks and Robot Platform:} For LHP, the task can be classified as observation (O), manipulation (M), and navigation (N). Depending on whether the observation task is required to actively acquire information, it can be divided into passive observation (PO) and active perception (AP) \cite{bajcsy1988active,bajcsy2018revisiting}. 
The types of robot platforms and the consequent task capabilities depend on each other.  For tasks that require both navigation and manipulation ability, mobile manipulators have been widely deployed due to their high versatility \cite{mobile_mani_p&p,mittal2022articulated}. With the combination of active perception and manipulation, the hand-eye system \cite{hand-eye-survey} enables the robot to switch between near-range observation and the required manipulation tasks. 


\textbf{Scene Knowledge:}  Planning algorithms for scenarios with complete prior scene knowledge advances for tasks involving complex motion primitives \cite{Symbolic_mobile_manipulator} or long-action sequencing in multiple-object manipulation \cite{gao2023orla} (Tab.~\ref{tab:alg_compare}, row 2). Non-adaptive informative planning, such as sensor planning in \cite{agriculture_UAV_UGV}, assumes deterministic action outcomes and complete scene information for planning (Tab.~\ref{tab:alg_compare}, row 1). When prior scene knowledge is incomplete, planning must incorporate perception alongside navigation or manipulation, as the robot needs to collect information incrementally from the scene during the execution of the trajectory  \cite{garrett2020online, muguira2023visibility}.  For partially observed environments with occlusions, Garrett et al. \cite{garrett2020online} developed an online replanning algorithm based on TAMP. Their method employs a particle filter to represent object pose beliefs, reducing belief states to deterministic values through self-loop action determination (Tab.~\ref{tab:alg_compare}, row 4).  Contact-based planning in blindfolded traveler's problem \cite{saund2019blindfolded} updates the beliefs of obstacles while planning the best expected route (Tab.~\ref{tab:alg_compare}, row 3).

\textbf{Metrics and Optimality:}
Planning performance is often evaluated based on metrics such as time to complete a task, task success rate, trajectory length, etc. Depending on the decision variable, they can be classified as two types: i) task-level metrics refer to metrics to assess task assignment or action sequence as decision variables. For a given task plan, if changing symbolic action, adding / removing action, or changing the order of actions does not improve overall performance, the task plan is optimal \cite{lo2018petlon}. ii) motion/state-level optimality is defined on robot motion, trajectory, and poses. For navigation-type tasks, the focus of the algorithms is to find the shortest travel distance using approximation algorithms \cite{agriculture_UAV_UGV,saund2019blindfolded}. ORLA* \cite{gao2023orla} for the tabletop rearrangement problem find the sequence of pick-and-place (p\&p) actions that minimizes execution time.  TAMP often solves optimality at the continuous motion/state level \cite{clutter_grasping}, but does not optimize the efficiency of the plan at the discrete task/action level \cite{garrett2021integrated}. Although \cite{garrett2020online,muguira2023visibility} share observation and manipulation steps, their algorithms focus on the feasibility of completing a complex multistage task in a single scene instance. In contrast, our problem is focused on minimizing energy usage in long run average due to our application characteristics. Additionally, our focus is on the aspect of spatial task multiplexing rather than motion/state-level planning, as the latter is often straightforward in an open agricultural field.

RHTP extends the single target Coupled Active Perception and Manipulation problem (CAPM) introduced in \cite{xie2023coupled}.  While CAPM introduces the Observe-First-Manipulate-Later (OFML) constraint for single targets, RHTP advances this concept by addressing multiple clustered targets and incorporating spatial task multiplexing. This extension significantly broadens the applicability and efficiency of the approach in complex, multitarget environments.


\section{Background and Problem Formulation}


\noindent \textbf{Regenerative point:} As the robot repetitively performs weed removal tasks in the field, it periodically captures overhead images from the row center line by elevating its camera to provide an overhead view. We refer to weeds as targets. Due to the perspective limit, these images provide only a low-resolution scene description, containing the target regions of interest (TROI) depicted as colored circular regions in Fig.~\ref{fig:scene}. TROIs are typically identified through object detection algorithms applied to these overhead images. The planning algorithm needs to find an energy-efficient plan to process all known TROIs. The moment when an overhead photo is taken serves as the natural regenerative point in this renewal process. An instance of the RHTP problem follows:

\subsection{Assumptions}\label{ssc:assumption}
{
\begin{enumerate} 
\item[a.1] \label{assumption:holonomic}The mobile platform is holonomic. 
\item[a.2]\label{assp:no-obstacle}  All weed targets are reachable. 
\item[a.3]\label{assp:subsequence} To ensure stability, the robot does not move its base and arm simultaneously.  
\item[a.4] The robot arm is in a rest position while the base is moving. 
\item[a.5] The energy usage of arm movement for each observation / flaming manipulation task can be approximated as constants. \label{assp:arm-cost}
\end{enumerate}
}

Assumptions a.1 and a.2 are common in early stage weed removal applications where limited weed and crop heights do not impede robot mobility. In addition, quadruped robots ensure holonomic motion.  Assumptions a.3 and a.4 enhance stability and simplify obstacle avoidance.  Assumption a.5 is valid for repetitive tasks.  


\subsection{Scene Description}
Denote the robot workspace $\mathcal{W} \subset \mathbb{R}^3$ with the $x$-$y$ plane overlap with the horizontal ground plane and the $z$ axis pointing upward. Given $n$ targets with index set $\mathcal{I} = \{1,...,n\}$, TROI and manipulation point of interest (MPOI) for the target $i\in \mathcal{I}$ are defined as follows. 
\begin{itemize}

    \item[$T_i$]Target $i$'s TROI defined by the geometric center at $X^c_{i}=[x_{i},y_{i},0] \in \mathcal{W}$ of the target and its radius $r_i$. $T_i := \{X=[x,y,z]:  \|X-X^c_{i}\|_2^2 \le r^2_{i}, z\ge 0\} \subset \mathcal{W}$. 

    \item[$X_i$]Target $i$'s MPOI, $X_i \in T_i$. Note that TROI's geometric center $X^c_{i}$ does not necessarily overlap with $X_i$. 

\end{itemize}

For each target $i$, a reasonable model of the distribution is a uniform distribution over $T_i$ since $X_i$ could be anywhere in $T_i$. However, the weed growth point is likely to be located in the center of $T_i$ due to biological symmetry. We model $X_i$ as a truncated normal distribution: $X_i\sim\mathcal{N}_T(X^c_i, \Sigma_i, r_i)$, $1\leq i \leq n$ where $X^c_i$ is the center of $T_i$, $\Sigma_i$ is the covariance matrix, and $r_i$ is the truncation radius. The initial belief space of the $n$ MPOIs is thus defined as:
\begin{Def}{(MPOI Belief Space)} \label{Def:belief}
For a vector of $n$ MPOIs,  $\mathbf{X} = [X_1,...,X_n]$, its belief space $\mathcal{B}_\mathbf{X}$ can be represented as $\mathbf{X} \sim \mathcal{B}_\mathbf{X}$ with
\begin{equation}\label{eq:belief_init}
    \mathcal{B}_\mathbf{X} = [\mathcal{N}_T(X^c_1, \Sigma_1), ..., \mathcal{N}_T(X^c_n, \Sigma_n)]. 
\end{equation}
\end{Def}
Each close-up observation collapses the corresponding $i$-th belief $\mathcal{N}_T(X^c_i, \Sigma_i,r_i)$ into deterministic values.


\subsection{Problem Formulation}
RHTP with mobile manipulators for a set of weed-flaming tasks is the base placement sequence selection and the task assignment problem. Define $\mathcal{K}=\{0,...,K\}$ as the time sequence index set. $\mathbf{x}_{b,k} \in \mathcal{X}_b$ denotes base pose at the $k$-$th$ stop. The task plan can be represented as the sequence of base stops and the target index set assigned to the $k$-$th$ base stop: $\pi = \{(\mathbf{x}_{b,k}, \mathcal{I}_{k} )\}_{0:K}$, where $\mathcal{I}_{k}\subset \mathcal{I}$.

\subsubsection{Observation/Manipulation Task Constraints} \label{sec:constraints}

For our mobile manipulator with hand-eye system, the observation reachability constraint forces the arm to approach the target $i$ so that $T_i$ is fully covered in the camera image with sufficient resolution to identify $X_i$. The manipulation reachability constraint determines whether the manipulator can reach the target while avoiding self-collision or an inability to perform the weed-flaming task without damaging the robot. Given a known manipulator configuration, we model the observation/manipulation reachability constraint as a binary function with $\mathds{1}_{\mbox{\tiny O/M}}(\mathbf{x}_{b,k}, T_i,\bar{X}_{i}) = 1$ which means that the constraint is satisfied. The detailed formulation will be explained later. 

Another type of constraint is observe-first-and-manipulate-later (OFML) temporal constraint that forces the order of the two different tasks:
\begin{Def}{(OFML)}\label{def:OFML}
For the target $i$, let $k_{\tau}, k_{\iota} \in \mathcal{K}$  be time indices for its observation and manipulation actions in $\pi$, respectively. OFML can be described as $k_{\tau} \le k_{\iota}, \forall i \in \mathcal{I}$. 
\end{Def}

\subsubsection{Energy cost}


We evaluated plans $\pi$ based on energy consumption due to limitations of battery capacity onboard. Our goal is to compare the energy costs of different task sequences instead of computing the precise execution costs.  Given Assumption a.5, the arm energy does not vary with the task sequence. Therefore, we focus on the energy consumption from the robot base movement. We follow the
energy modeling in \cite{hou2018energy} using a combination of a fixed initial starting cost for the base and a variable energy cost,  but other energy cost models can also be used. Denote $e_d(\mathbf{x}_{b,k-1},\mathbf{x}_{b,k})$ as the obstacle-free travel distance between $\mathbf{x}_{b,k-1}$ and $\mathbf{x}_{b,k}$, for a plan sequence, the normalized cost is 
\begin{equation}\label{eq:cost}
c(\pi) = K  + \gamma \sum_{k=1}^{K}e_d(\mathbf{x}_{b,k-1},\mathbf{x}_{b,k}). 
\end{equation}
where $\gamma$ is the energy cost coefficient, a ratio between the starting cost and the variable costs of the base, $\gamma > 0$.

\begin{Pro}[RHTP] \label{problem_def}   
 Given the start state $\mathbf{x}_s$, goal state $\mathbf{x}_g$, and a set of TROIs $\{T_i:i \in \mathcal{I}\}$, sequentially find a task plan $\pi$ and for action and key states to obtain MPOI set $\{X_{i}:i \in \mathcal{I}\}$ to guide and execute the subsequent manipulation task with the minimum expected energy cost:
\begin{align}
\min_{\mathbf{a}, \mathbf{x}} \quad &  \mathbb{E}_{\mathbf{X} \sim \mathcal{B}_\mathbf{X}}[c(\pi)] \label{eq:obj-CAPM}\\
\textrm{s.t.} \quad & \bigwedge_i \mathds{1}_{\mbox{\tiny O/M}}(\mathbf{x}_{b,k}, T_i,\bar{X}_{i}) = 1 , \forall i \in \mathcal{I}, \label{eq:task_const}\\
& \mathbf{x}_{b,0} = \mathbf{x}_s,  \mathbf{x}_{b,k_{\mbox{\tiny max}}} = \mathbf{x}_g, \mbox{and OFML in Def. \ref{def:OFML}} 
\end{align}
\end{Pro}


\section{Algorithm} \label{sec:alg}

An effective way to reduce energy is to group the targets using clustering. Two targets are considered to be in the same cluster if there exist mobile manipulator configurations that can observe/manipulate either target without moving the base. As an example, Fig.~\ref{fig:scene} colors the clusters differently. Clustering enables spatial task multiplexing (STM) of robot base motions. It is clear that STM can reduce base movements, reduce energy usage, and increase speed. 
   
Meanwhile, the OFML constraint in Def.~\ref{def:OFML} prevents deterministic planning before MPOI is observed. We need to employ the MPOI belief space in Def.~\ref{Def:belief} to construct the task sequence. For each cluster, a wise strategy is to increase the likelihood of spatial multiplexing by choosing the optimal base positions. As a result, there is a combinatorial nature to this problem in the iterative process. We begin with introducing the probabilistic target reachability map (PTRM).




\subsection{Probabilistic Target Reachability Map}

The construction of the PTRM involves converting the MPOI belief $\mathcal{B}_\mathbf{X}$ into the probability that the mobile manipulator can reach a target from a given base pose $\mathbf{x}_b \in \mathcal{X}_b$, as defined in \eqref{eq:task_const}. This constraint is inherently probabilistic due to the random distribution of unobserved targets.


The robot base state constraints can be constructed through the circular approximation \cite{sandakalum2022motion} by inverse kinematics of arm configurations. The reachable space for the robot base $\mathcal{R}_o(T_i) \subset \mathcal{X}_b$ to observe $T_i$ can be approximated to a compact, closed donut-shaped region due to camera coverage and resolution constraints. Similarly, given $X_i$, the reachable space $\mathcal{R}_m(X_i) \subset \mathcal{X}_b$ for the robot base enabling manipulation is also a closed donut-shaped region due to self-collision and kinematic constraints. Detailed derivations can be found in~\cite{xie2023coupled}. Eq.~\eqref{eq:task_const} is satisfied if $\mathbf{x}_b \in \mathcal{R}_o(T_i)$ and $\mathbf{x}_b \in \mathcal{R}_m(X_i)$ when observing $X_i$. In general cases, the feasible manipulation region $\mathcal{R}_m(X_i)$ is smaller than the feasible observation region $\mathcal{R}_o(T_i)$  because the camera on board can often cover a much larger region than the arm can reach; as analyzed in \cite{xie2023coupled}. 

Considering that the probability on the estimated $\hat{X}_i$ is truncated and is limited in a small region of TROI, we further approximate the reachability constraints such that all the base states that satisfy the manipulation constraint also satisfy the observation constraint:
\begin{equation} \label{eq:ring_cond}
    \mathcal{R}_m(\hat{X}_i) \subset \mathcal{R}_o(T_i), \forall \hat{X}_i \in T_i.
\end{equation}
This approximation allows us to formulate $\mathcal{R}_m(\hat{X}_i)$ as the mutual base region for the joint observation/manipulation (O/M) task. This leads to the following new formulation of reachability constraint on the base state: 
\begin{equation} \label{eq:base_task_const}
    \mathds{1}^b_{\mbox{\tiny O/M}}(\mathbf{x}_{b},\hat{X}_{i}) =
\begin{cases}
1, ~~~\text{if}~  \mathbf{x}_b \in \mathcal{R}_m(\hat{X}_i)\\
0, ~~~~\text{Otherwise.}
\end{cases}
\end{equation}
Under the simplification, we replace
reachability constraint \eqref{eq:task_const} 
with \eqref{eq:base_task_const} for base placement in our RHTP problem.  

Due to the OFML constraint, the reachable region for the manipulation task is nondeterministic. We calculate the probability of performing a successful task in pose $\mathbf{x}_b \in \mathcal{X}_b$, based on the current belief space $\mathcal{B}_\mathbf{X}$. 
Next, we are ready to introduce the initialization and update of the PTRM.

\begin{figure}[htbp]
\centering
\vspace*{-0.1in}
\subfloat[\label{fig:scene_plot}]{\includegraphics[width=1.1in]{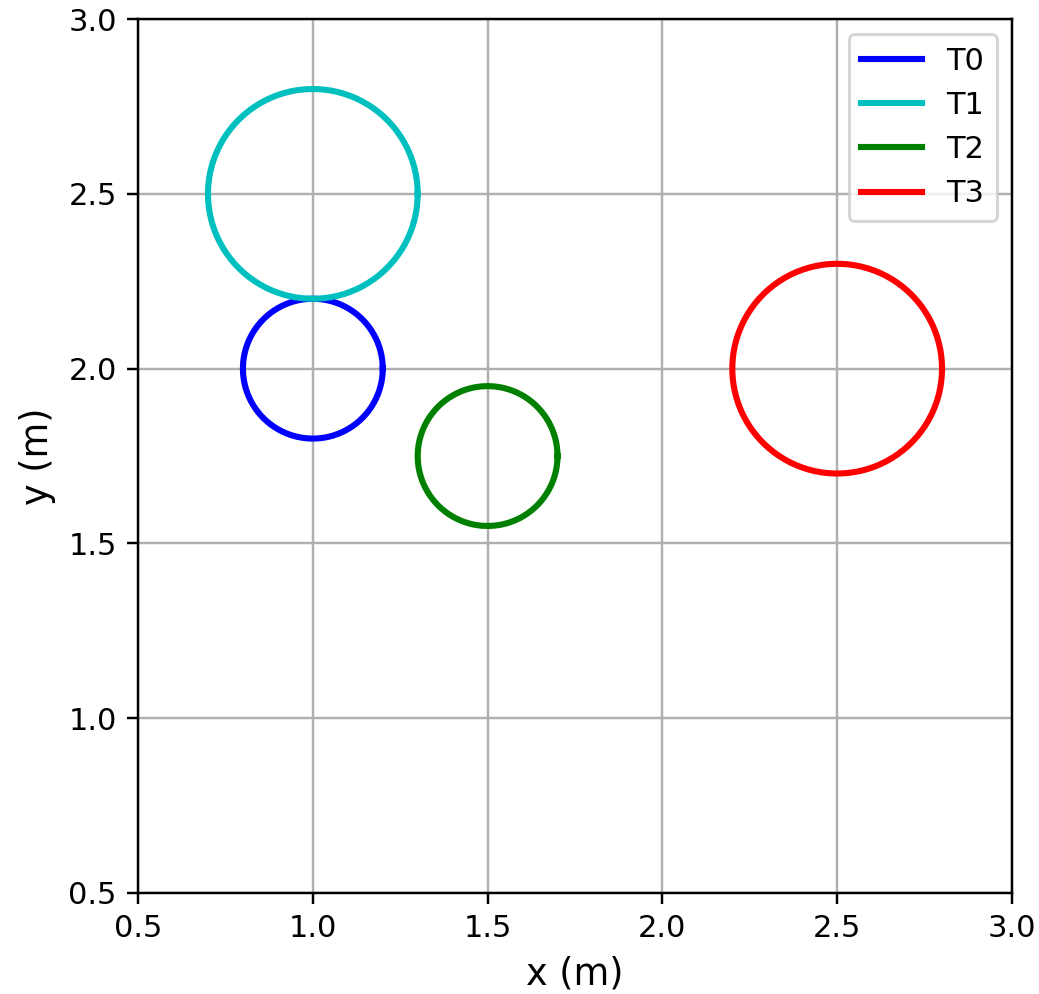}}
\hspace*{.05in}\subfloat[\label{fig:PTRM}]{    \includegraphics[width = 1.4in]{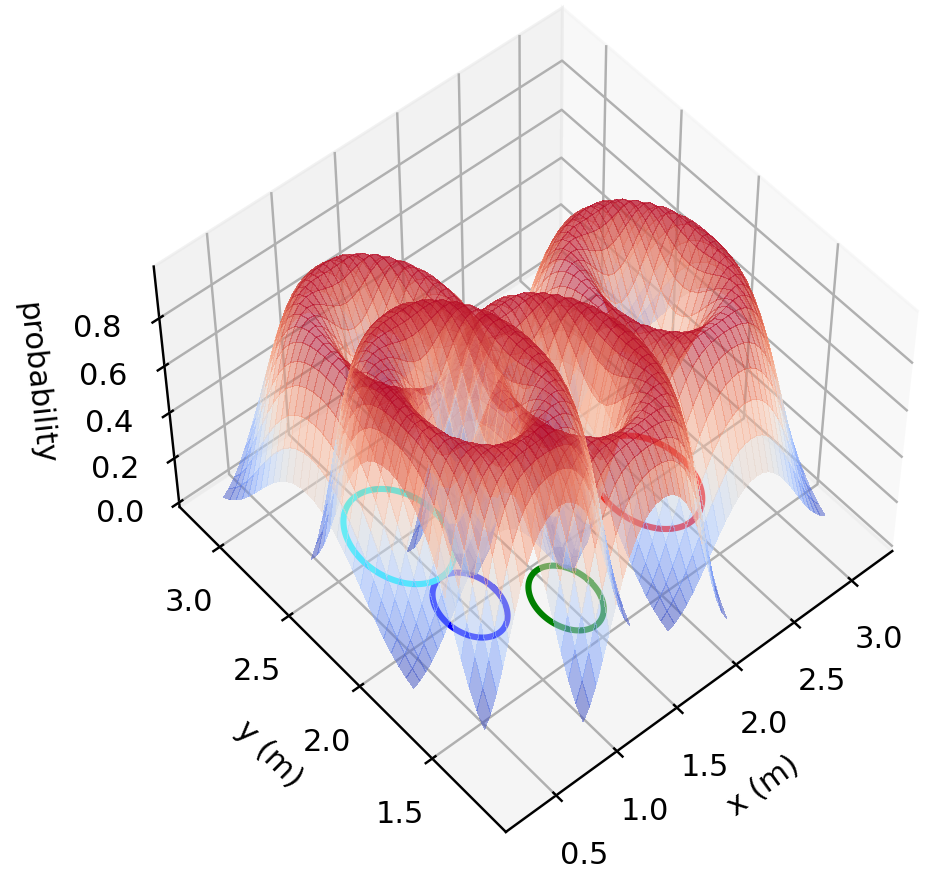}}
\caption{(a) A running example with 4 targets and the TROIs. (b) PTRM (2D at x-y plane). These probabilities are superimposed on one coordinate system.  
}
\vspace*{-0.2in}
\end{figure}

\vspace{.05in}\noindent\textbf{Initialization:}
at the starting point, the probability of base position $\mathbf{x}_{b}$ that allow the robot to reach the target $i$ to execute a successful task is,
\begin{align}
 \label{eq:init_prob}
p_i(\mathbf{x}_{b})  &:=  p(\mathbf{x}_{b} \in \mathcal{R}_m(\hat{X}_i)|\hat{X}_i) 
 \\ &=    \int f_T(\hat{X}_i; X^c_{i}, \Sigma_i, r_i) \mathds{1}^b_{\mbox{\tiny O/M}}(\mathbf{x}_b,\hat{X}_i)  d \hat{X}_i,   
\end{align} 
where $f_T(\cdot)$ is the probability density function of truncated normal distribution.
Since $p(\hat{X}_i)$ is a truncated probability, we define a base work space $S_i$ for target $i$ as base configuration set with non-zero probability to reach the target $i$ as $ S_i = \{\mathbf{x}_b : p_i(\mathbf{x}_b)  > 0\}$, 
which covers all potential robot base configurations with non-zero probability to complete the task. 

\vspace{.05in}\noindent\textbf{Update:} During run time, once the target $i$ is perceived and its MPOI is detected after robot movement, $X_i$ becomes deterministic, $p_i(\mathbf{x}_{b}) := 1, \forall \mathbf{x}_{b} \in \mathcal{R}_m(X_i).$ 
Consequently, the MPOI belief $\mathcal{B}_\mathbf{X}$ collapses to the observed deterministic value.  We also update the base task space $S_i = \mathcal{R}_m(X_i).$

\subsection{Task Space Partition} \label{sec:task_partition}
\begin{figure}
    \centering
    \includegraphics[width = 3.1in]{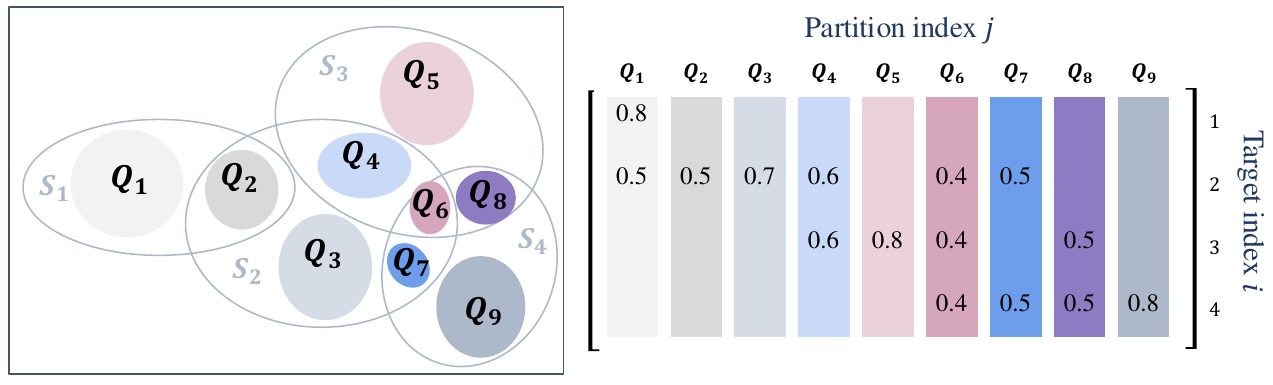}
    \caption{PTRM partition. Left: a topological visualization of task space partition. Right: partition's task success probabilities matrix. These probabilities are based on location and not normalized against either $i$ or $j$.}
    \label{fig:prob_table}
    \vspace*{-0.15in}
\end{figure}

The PTRM provides the sets of feasible base states to execute the O/M task for each target. The overlapping regions are the base pose set for the robot where it is possible for the robot to achieve STM. For example, the base state within the region $S_1 \cap S_2$ means that it is possible to multiplex the tasks for both targets if the robot is located in the region. To further investigate the possibilities of STM, we partition the base task space $\mathcal{S}$ into regions based on the task space intersection so that each region corresponds to a unique combination of STM (see Fig.~\ref{fig:prob_table}). 


\begin{Def}{(Task Space Partition)} \label{def:task_space_partition}
    A task space partition $\mathcal{Q} = \{Q_j : j\in \mathcal{I}_Q \}$ is constructed from a collection of the base task space sets $\mathcal{S} = \{S_i : i\in \mathcal{I}\}$ such that 
\begin{itemize}
    \item[(a)] $\bigcup_{j\in \mathcal{I}_Q} Q_j = \bigcup_{i\in \mathcal{I}} S_i$,\\\vspace*{-0.1in}
    \item[(b)] If $j \neq l $ then $ Q_{j}\cap Q_{l} =  \emptyset $, and \\\vspace*{-0.1in}
    \item[(c)] If $Q_j \subseteq S_i$, we label $S_i$ as the parent set of $Q_j$. For each partition $Q_j$, it may have multiple parents. Its parent index set is denoted as $\mathcal{I}_j \subset \mathcal{I}$. For two partitions $Q_j$ and $Q_l$, if $j \neq l $, then $ \mathcal{I}_j \neq \mathcal{I}_l$.
\end{itemize}
\end{Def}
Conditions (a) and (b) in Def.~\ref{def:task_space_partition} ensure that $\mathcal{Q}$ is a partition. Condition (c) states that each partition $Q_j$ corresponds to a unique STM combination.  An example partition is shown in Fig.~\ref{fig:prob_table}. For each partition of the task space $Q_j$, we compute the likelihood that the reachability constraint of the task $i$' is satisfied when the robot base is placed there. Therefore, we define the probability matrix to capture each partition's task success probabilities $\mathbf{P}$ with each element $\mathbf{P}_{i,j}$ calculating the probability of target $i$ can be successfully reached from any base state from the partition region $Q_j$, i.e. $\mathbf{x}_b \in Q_j $: 
\begin{equation}
    \mathbf{P}_{i,j} := \frac{\int_{\mathbf{x}_b \in Q_j} p_i(\mathbf{x}_b) d \mathbf{x}_b}{\int_{\mathbf{x}_b \in Q_j} 1 d \mathbf{x}_b}  
\end{equation}
With each $\mathbf{P}_{i,j}$ computed for all partition-target combinations, we obtain the matrix $\mathbf{P}$ to represent the probability of each STM combination. $\mathbf{P}$ is updated each time the PTRM update occurs. The right side of Fig.~\ref{fig:prob_table} shows a partition's task success probabilities matrix example.

\subsection{Base Placement Planning}\label{ssc:base-planning}

Recall the objective function in \eqref{eq:obj-CAPM}, we want to minimize the energy $\mathbb{E}[c(\pi)]$. Obviously, we can condition the expected energy on the selected task space partitions which can be viewed as high-level decision variables that direct the robot to stop and execute tasks. 

We begin with the probability of reaching the target $i$ for a given partition set. Define the task space variable $\boldsymbol \phi = [\phi_1, …, \phi_r]$ corresponding to the partition index set $\mathcal{I}_Q = \{1,...,r \}$, where binary variable $\phi_j = 1$ means that the partition $Q_j$ is selected and $0$ means otherwise. 
Let $M_i$ be the event in which the robot successfully completes the task for the target $i$. Then we have conditional probability $p(M_i|\boldsymbol \phi)$ derived from the joint failure probability
$p(M_i|\boldsymbol \phi) = 1- p(\bar{M}_i|\phi) = 1-\prod_{j}(1-\phi_j\mathbf{P}_{i,j})$.
Note that $\boldsymbol\phi$ contains regions that the robot base needs to stop and perform tasks and represents a high-level task plan that corresponds to the output $\pi$. 

Recalling the energy cost function in \eqref{eq:cost}, we need to minimize the number of selected regions, which reduces the fixed starting cost, since the robot stops at each selected partition. Meanwhile, the distance traveled must be minimized. The RHTP can be viewed as a set cover problem with symmetric traveling salesperson problem (TSP) on the selected task space partition set.  

Here, we define the distance traveled between two partition regions $Q_j$ and $Q_l$ as the minimum variable energy cost $e_d(Q_j,Q_l) = \min_{\mathbf{x}_j \in Q_j, \mathbf{x}_l\in Q_l } e_d(\mathbf{x}_j, \mathbf{x}_l)$. We reformulate the cost function to incorporate set selection as decision variables, accounting for both the base stop times and the energy cost of traveling between selected regions from the initial to the final states with a weighting coefficient $\gamma$, as shown in Eq.~\eqref{eq:bpr_obj}. 

Let us define the vertex set as the union of the region index set $\mathcal{I}_Q$ with a size of $r$ and the start state with index $0$ / end state index with index $r+1$ such that $V = \mathcal{I}_Q \cup \{0, r+1\} = \{0,1,...,r,r+1\}$. $Q_0$ and $Q_{r+1}$ are sets that contain only the start and end poses, respectively, which are also source and sink nodes of the TSP tour on the graph if we treat each partition/region as a node. We define the binary path selection variables $\boldsymbol \xi$ where $\xi_{j,k}  = 1 $ represents the region $k$ visited just after the region $j$ and $0$ otherwise. Since $e_d(Q_j,Q_k) = e_d(Q_k,Q_j)$, the TSP is symmetric so that $\xi_{j,k}$ and $\xi_{k,j}$ have the same meaning. The task space partition selection problem 
is formulated as follows, 
{
\fontsize{9}{10}\selectfont
\begin{align} 
\min_{\boldsymbol  \phi, \boldsymbol \xi} \quad &   \mathbf{\phi}^\mathsf{T}\mathbf{\phi} + \gamma \sum^{r+1}_{j = 0} \sum^{r+1}_{k = j+1} \xi_{j,k} e_d(Q_k,Q_j) \label{eq:bpr_obj} \\
\textrm{s.t.} \quad &   p(M_i|\boldsymbol \phi) \ge \delta, \hspace{9em} \forall i\in \mathcal{I}, \label{eq:bpr_prob_constraint}  \\
 &  \sum^{j-1}_{k=0} \xi_{k,j} +\sum^{r+1}_{k=j+1} \xi_{j,k} = 2\phi_j, \hspace{2em} \forall j \in \mathcal{I}_Q, \label{eq:path_constraint} \\
& \sum_{j\in \mathcal{I}_Q} \xi_{0,j} =1,  \sum_{j\in \mathcal{I}_Q} \xi_{j,r+1} =1, \label{eq:s/e_constraint} \\
&  \sum_{j\in S}\left( \sum_{\substack{k\in \mathcal{I}_Q \setminus  S \\ k<j}} \xi_{k,j} +\sum_{\substack{
k\in \mathcal{I}_Q \setminus S \\ k>j }} \xi_{j,k}  \right) \le 2, \nonumber\\ 
&\hspace{10.2em}\forall S\subset \mathcal{I}_Q, |S|\ge 2. \label{eq:subtour_constraint}  
\end{align}
}
By choosing a proper probability threshold $\delta$, the constraint \eqref{eq:bpr_prob_constraint} ensures that the partitions of the selected task space have a high probability to complete all tasks. Motivated by the Dantzig-Fulkerson-Johnson formulation of the TSP, the constraints \eqref{eq:path_constraint}, \eqref{eq:s/e_constraint}, and \eqref{eq:subtour_constraint} ensure a complete tour from the start state to the end state that passes through the selected task space partitions. Eq. \eqref{eq:path_constraint} requires that two paths be connected for each selected region.  Eq. \eqref{eq:s/e_constraint} ensures that the start and end nodes should be connected to the path. Eq. \eqref{eq:subtour_constraint} is the subtour elimination constraint. 

A solution to this MINLP problem can be obtained using
Branch-and-Bound solvers \cite{borchers1994improved,Gentilini2008}. The optimization results are the selection of partitions of the task space indexed by non-zero entries in $\boldsymbol{\phi}$ and a visit sequence between these partitions specified by the binary path selection variables $\boldsymbol \xi$. Through the partition selection variable, we know that the number of stop times for the base is $\kappa = \boldsymbol{\phi}^T\boldsymbol{\phi}$. The sequence of partition ($\mathbf{q}$) to be visited can be generated from the path variable $\boldsymbol \xi$ as an index list. For the running example, $\mathbf{q}=[0, 2,8,10]$ from Fig.~\ref{fig:prob_table}. 


The final step is to generate a task plan $\pi$ by finding the base state $\mathbf{x}_b$ of each region in $\boldsymbol \xi$ that defines the shortest path in the same order of region selection. Solving the optimal solution $\mathbf{x}_b$ is trivial and we skip the details here for brevity.

\section{Experiments}

\subsection{Experiment Setup}
We have implemented the proposed RHTP algorithm using Python 3.8.10 on a Ubuntu 22.04 PC machine with the Intel(R) Core\texttrademark\  i7-10700K CPU @ 3.80GHz. We have tested the algorithm based on field data collected from our custom mobile manipulator (see Fig.~\ref{fig:scene}) to validate the performance for the real application. 

\begin{figure}[ht!]
    \centering
    \includegraphics[width =3.3in]{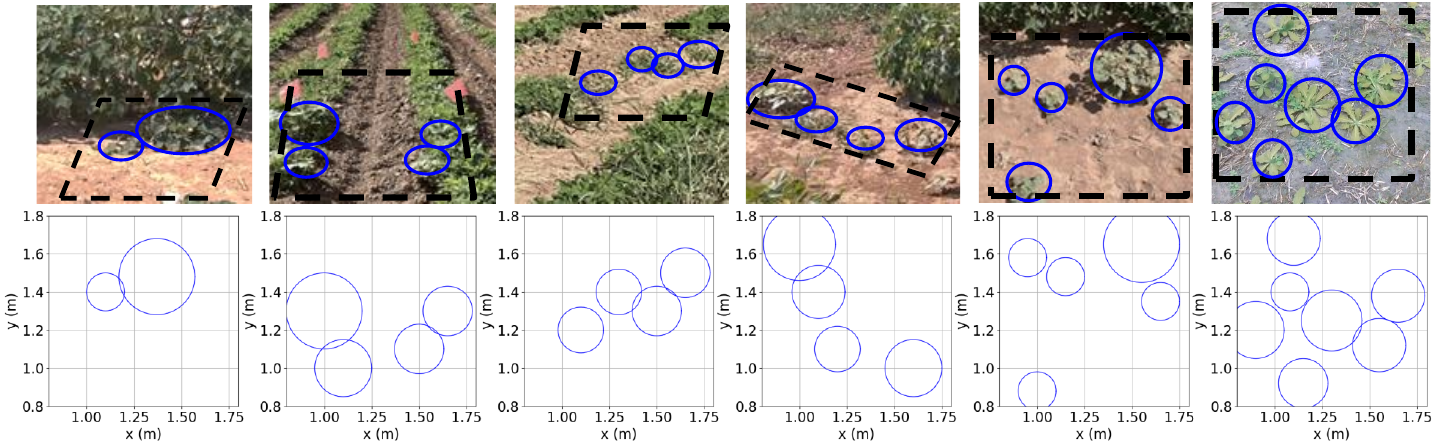}
    \caption{Sample scene configurations with different target densities from early/late grow stage in cotton field and test bed. }
    \label{fig:scene_density}
    \vspace*{-.1in}
\end{figure}
\noindent\textbf{Task and System Settings:}
We have mounted a 6 degree-of-freedom (DoF) Unitree Z1\texttrademark\ manipulator on the Boston Dynamics Spot Mini\texttrademark\ quadruped robot for the weed removal task. The Spot Mini\texttrademark\ carries all attached accessories (i.e. an additional computer, flammable fuel container, igniter, fuel pipes, relay controlled valves, control circuit, and a Unitree Z1 arm \cite{wang2024preciseroboticweedflaming}). 
We deployed the mobile manipulator in cotton fields to collect data to initialize the simulation used to evaluate our algorithms as described in \cite{wang2024preciseroboticweedflaming}. 


\noindent\textbf{Test Scene Configurations:}
The test scene is obtained from images collected from the cotton field at different stages of weed growth, as shown in Fig.~\ref{fig:scene_density}.  To validate the performance of the algorithm in different scenarios, we define the target density as the ratio between the number of weeds $| \mathcal{I}| $ and the area of the ground plane of the workspace $\rho = \frac{ | \mathcal{I}| }{\mbox{Area}(\mathcal{W}^{g})} $. We have tested a total of 50 scene instances, where for each instance $\mbox{Area}(\mathcal{W}^{g}) = 1\ \mbox{m}^2$ and the density of the weed varies from 1 to 7. 

\noindent\textbf{Baseline Algorithm:}
The baseline algorithm for comparison is a greedy naive algorithm that extends from the single target CAPM \cite{xie2023coupled} algorithm simply by handling one target at a time until all targets are handled. The target chosen at each step is the closest target that has not been processed. We call this algorithm Naive-CAPM. This can be considered as algorithms based on the hierarchy planning TAMP.

\subsection{Result}
\subsubsection{Speed Test} We have measured the run time of our algorithm, which is $2.2$ seconds on average. Since the execution time of actions (base, arm movements, and flaming) is much longer than the planning time, the run time of the algorithm is not the bottleneck in the weed removal application, because the computation can be done while the robot is in motion finishing the previous step.

\subsubsection{Offline Performance Comparison}
We have compared our proposed algorithm (RHTP) with Naive-CAPM for energy usage according to \eqref{eq:cost} and the number of replans which is the number of times iteration is needed to complete all tasks. 
For each instance, we sample 1000 possible combinations of MPOIs for the initialization of belief according to \eqref{eq:belief_init} where the initial belief parameter is $\Sigma_i =  r_i \mathbf{I}_{2\times2}$ and compute the average result for all metrics. The energy cost and the RHTP parameters are set as $\gamma = 1.12 $ and $\delta =0.7$ based on the hardware testing of our robot, respectively. 

First, we compare the two algorithms using all 50 instances in the test data. To separate the effect of density from the uncertainty introduced by prior observation (reflected through the radius), we set the TROI with the same radius. 


\begin{figure}[]
    \centering
    \includegraphics[width=3.4in]{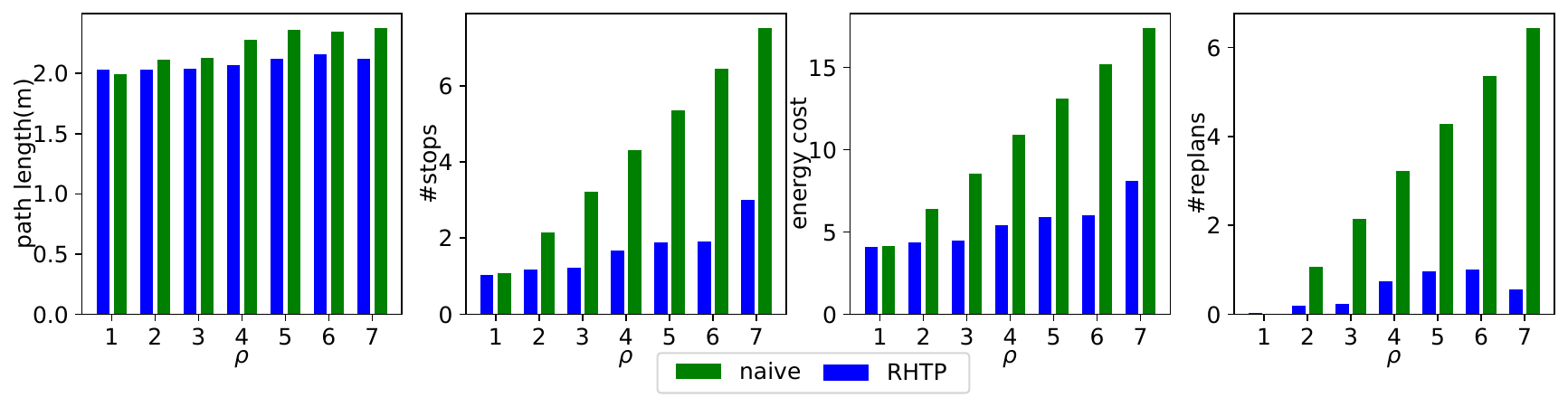}
    \caption{Performance comparison between the Naive-CAPM and the RHTP algorithms.}
    \label{fig:density-exp}
    \vspace*{-.2in}
\end{figure}

The plot in Fig.~\ref{fig:density-exp} shows the result. When the target density is low $\rho = 1$, the Naive-CAPM and RHTP algorithms perform very similar to each other. This is expected because if there is no cluster, they are essentially the same algorithms. As the target density increases, the RHTP has a shorter path length and less number of stops compared to the Naive-CAPM (result trajectory samples are shown in the video attachment).  As for the number of replans, the RHTP can achieve less than 1 replan across the board. For the Naive-CAPM method, since one iteration can only find the position for one stop, and the task plan requires the robot to stop for every target, the replan number increases with the density of target increase. In all metrics, the RHTP algorithm outperforms the Naive-CAPM algorithm more significantly as the density increases further. This is very desirable because our RHTP algorithm is more efficient when needed.

\begin{figure}[hbpt!]
    \centering    \vspace*{-.1in}
    \includegraphics[width=3.4in]{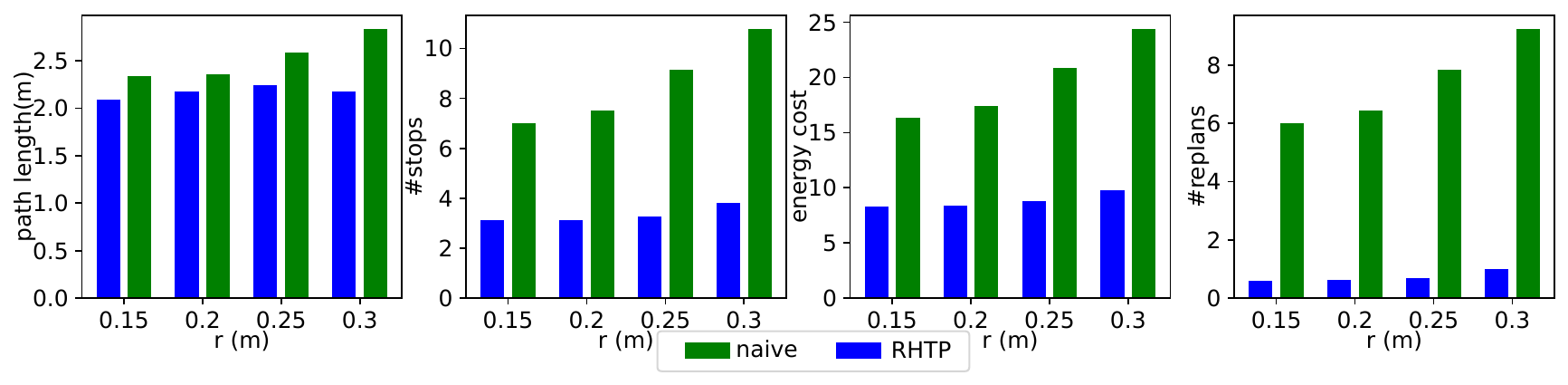}
    \caption{Performance comparison between the Naive-CAPM and the RHTP at different TROI radius settings. The horizontal axes are TROI radius.}
    \vspace*{-.2in}
    \label{fig:radius-exp}
\end{figure}

To validate the effect of the TROI radius, we choose the instances with the same density $\rho =5 $, where we have the most instance at this density (15 instances). We run both algorithms on this group of instances. We set the radius of the region to $0.15,0.20,0.25,0.30$ meters, respectively. As shown in Fig.~\ref{fig:radius-exp}, both algorithms are affected by the increase in region size because the increase in region size means a greater uncertainty of perception in prescan. In comparison, it is clear that RHTP still outperforms Naive-CAPM in all metrics. The advantage is more significant when the TROI radius (i.e. perception uncertainty) grows larger.

\section{Conclusion and Future Work}
We proposed RHTP algorithm to enable the mobile manipulator to perform repetitive observation and manipulation tasks efficiently in precision agriculture applications. The RHTP algorithm was designed to reduce the long-term average energy consumption rate by minimizing the expected energy use. We reduced the optimization problem to a region-based set-coverage problem by conditioning all possible spatial task multiplexing possibilities in different regions. In experiments, the algorithm demonstrated a significant improvement over its counterpart in all tasks. Moreover, the improvement is more significant when the target density increases, which is desirable. Future work will focus on improving base and arm motion coordination, and developing multi-robot algorithms for repetitive tasks. 

\section*{Acknowledgement}
{\small The authors thank D. Shell, J. O'Kane, S. Darhba, A. Jiang, W. Wang, D. Wang, F. Guo, A. Kingery for their inputs and feedbacks.}

\bibliographystyle{ieeetr}
\bibliography{syxie}

\end{document}